\documentclass[runningheads]{llncs}
\usepackage{graphicx}
\usepackage{spverbatim}
\usepackage{paralist} 
\usepackage{url} 

\usepackage[frozencache=true,cachedir=minted-cache]{minted} 
\usepackage{listings} 
\usepackage{tabularx} 
\usepackage{cleveref}
\usepackage{nameref} 
\usepackage{graphicx}
\graphicspath{ {./img/} }
\usepackage[flushleft]{threeparttable} 
\usepackage{svg} 
\usepackage[strict]{changepage}
\usepackage{framed} 

\definecolor{formalshade}{rgb}{0.95,0.95,1}
\definecolor{darkblue}{rgb}{0.0, 0.0, 0.55}

\newcounter{promptnumber}
\newenvironment{prompt}{%
 \refstepcounter{promptnumber}%
  \MakeFramed{\advance\hsize-\width\FrameRestore}%
  \noindent\hspace{-4.55pt}
  \begin{adjustwidth}{}{7pt}%
  \vspace{2pt}\vspace{2pt}%
   \textbf{Prompt \thepromptnumber:}~
}
{%
  \vspace{2pt}\end{adjustwidth}\endMakeFramed%
}

\begin{document}

\title{LLM-assisted Knowledge Graph Engineering: Experiments with ChatGPT}


\author{Lars-Peter~Meyer \inst{1,2,3}\orcidID{0000-0001-5260-5181} 
\and Claus~Stadler \inst{1,2,3}\orcidID{0000-0001-9948-6458} 
\and Johannes~Frey \inst{1,2,3}\orcidID{0000-0003-3127-0815}
\and Norman~Radtke \inst{1,2}\orcidID{0000-0001-9155-8920}
\and Kurt~Junghanns \inst{1,2}\orcidID{0000-0003-1337-2770} 
\and Roy~Meissner \inst{2,3}\orcidID{0000-0003-4193-8209}
\and Gordian~Dziwis \inst{1}\orcidID{0000-0002-9592-418X}
\and Kirill~Bulert \inst{1,2}\orcidID{0000-0002-1459-3754}
\and Michael~Martin \inst{1,2}\orcidID{0000-0003-0762-8688}}
%
\authorrunning{L.-P. Meyer et al.}
%

\institute{
  InfAI e.V. Leipzig, Germany,
  \email{lpmeyer@infai.org},
  \url{https://www.infai.org}
  \and
  AKSW research group, \url{https://aksw.org}
  \and
  Leipzig University, Germany, \url{https://www.uni-leipzig.de}
}

\maketitle              
\begin{abstract}
Knowledge Graphs (KG) provide us with a structured, flexible, transparent, cross-system, and collaborative way of organizing our knowledge and data across various domains in society and industrial as well as scientific disciplines. 
KGs surpass any other form of representation in terms of effectiveness. 
However, Knowledge Graph Engineering (KGE) requires in-depth experiences of graph structures, web technologies, existing models and vocabularies, rule sets, logic, as well as best practices. 
It also demands a significant amount of work.

Considering the advancements in large language models (LLMs) and their interfaces and applications in recent years, we have conducted comprehensive experiments with ChatGPT to explore its potential in supporting KGE. 
In this paper, we present a selection of these experiments and their results to demonstrate how ChatGPT can assist us in the development and management of KGs.

\end{abstract}
\section{Introduction}
In the last years, Artificial Intelligence (AI) has shown great promise in improving or revolutionizing various fields of research and practice, including knowledge engineering. 
The recent big leap in AI-based assistant chatbots, like ChatGPT (Generative Pre-trained Transformer) model, has created new opportunities to automate knowledge engineering tasks and reduce the workload on human experts. 
With the growing volume of information in different fields, the need for scalable and efficient methods to manage and extract knowledge from data that also adapt to new sources is critical. 
Despite the advances in research w.r.t. (semi)automation, knowledge engineering tasks still rely vastly on human experts. 
On one hand, this process can be time-consuming, resource-intensive, and susceptible to errors.
On the other hand, the reliance on human expertise in knowledge engineering exposes it to workforce shortages (as knowledge engineers are scarce and the demand is growing) and the risk of expertise loss. 
These factors can impact the resilience and sustainability of systems and operations that rely on knowledge engineering.
AI-based assistant bot approaches, such as ChatGPT, could bridge this gap by providing a unified tool for tasks in knowledge engineering, to reduce the workload of knowledge engineers themselves, but also make knowledge engineering more accessible to a broader audience.
ChatGPT, in particular, has shown promise in generating responses in a variety of syntactical representations (including code and markup languages) to user queries or task descriptions written in natural language. 

In this paper, we discuss and investigate the potential of ChatGPT to support or automate various knowledge engineering tasks (e.g. ontology generation, SPARQL query generation).
We will explore the benefits, pitfalls and challenges of using it and identify potential avenues for future research.

\section{Related Work}

\emph{ChatGPT}, a Large Language Model (LLM) published by OpenAI\footnote{\url{https://openai.com/blog/chatgpt}}, raised the interest in the broad field of Machine Learning (ML)\footnote{\url{https://aiindex.stanford.edu/wp-content/uploads/2023/04/HAI_AI-Index-Report_2023.pdf}} and especially LLMs\cite{Haque2022Ithinkthis} on a broad scale.
While there are current discussions and analysis on the capabilities of LLMs like ChatGPT in general (e.g. \cite{Bubeck2023SparksArtificialGeneral}), there is little in the area of knowledge graph engineering.
Ekaputra et al.\cite{Ekaputra2023DescribingOrganizingSemantic} gives a general overview of current research on the combination of the broad field of ML and semantic web.

Searching Google Scholar and Semantic Scholar with "knowledge graph ChatGPT", "ontology ChatGPT" and "rdf ChatGPT" in the beginning of April 2023 results in only two relevant papers.
The first one, \cite{Omar2023ChatGPTvsTraditional}, reviews the differences between conversational AI models, prominent ChatGPT, and state-of-the-art question-answering systems for knowledge graphs.
In their survey and experiments, they detect capabilities of their used frameworks but highlight ChatGPTs explainability and robustness.
The second one, \cite{Lin2023ContextBasedOntology}, discusses the usage of ChatGPT for database management tasks when tabular schema is expressed in a natural language.
They conclude among others that ChatGPT is able to assist in complex semantic integration and table joins to simplify database management and enhance productivity.
The applied approaches and results of these two papers indicate that the idea of using LLMs like ChatGPT in the field of KG engineering is encouraging and that the LLMs might assist KG engineers in their workflows.
Still, the research on the usage of LLMs for knowledge graph engineers is scarce and seems to be a new research area.

There exist some non- and semi-scientific resources which render the topic from a practical and experience perspective.
We want to highlight here a helpful blog post by Kurt Cagle \cite{Cagl2023GptTricks} on ChatGPT for "knowledge graph workers" and a blog post by Konrad Kaliciński \cite{Kalicinski2023Neo4jGpt} on knowledgegraph generation in Neo4J assisted by ChatGPT.

\section{LLM-Assisted Knowledge Graph Engineering - Potential Application Areas}\label{usecases}

In discussion rounds with knowledge graph engineering experts we identified the following preliminary list of potential use cases in the domain of knowledge graph engineering applicable to LLMs assistance:
\begin{itemize}
    \item Assistance in knowledge graph usage:
    \begin{itemize}
        \item Generate SPARQL queries from natural language questions (related experiment in \Cref{sec:ex:sparql-small-custom} and \Cref{sec:ex:sparql-mondial})
        \item Exploration and summarization of existing knowledge graphs (related experiment in \Cref{sec:ex:KG-exploration})
        \item Conversion of competency questions to SPARQL queries
        \item Code generation or configuration of tool(chain)s for data pipelines
    \end{itemize}
    \item Assistance in knowledge graph construction
    \begin{itemize}
        \item Populating knowledge graphs (related experiment in \Cref{sec:ex:fact-sheets-KG}) and vice versa    
        \item Creation or enrichment of knowledge graph schemas / ontologies
        \item Get hints for problematic graph design by analysing ChatGPT usages problems with a knowledge graph
        \item Semantic search for concepts or properties defined in other already existing knowledge graphs
        \item Creation and adjustment of knowledge graphs based on competency questions
    \end{itemize} 
\end{itemize}

Given the limited space of this paper, we evaluate a subset of the application areas with experiments in the following section.

\section{Experiments}
\label{sec:experiments}
To evaluate the capabilities of LLMs at the example of ChatGPT for assisting with knowledge graph engineering, we present several experiments and their results.
Further details about them is given in the \nameref{sec:onlineressources}.
Most experiments were conducted with ChatGPT with the LLM GPT-3.5-turbo\footnote{\url{https://platform.openai.com/docs/models/gpt-3-5}\label{fn:gpt-3-5-doc}} (named \emph{ChatGPT-3} from here on), some additionally with ChatGPT with the LLM GPT-4\footnote{\url{https://platform.openai.com/docs/models/gpt-4}\label{fn:gpt-4-doc}} (named \emph{ChatGPT-4} from here on).

\subsection{SPARQL Query Generation for a Custom Small Knowledge Graph }
\label{sec:ex:sparql-small-custom}
For a first evaluation, we designed a small knowledge graph as shown in Listing \ref{lst:example-kg}.
Specifically, we wanted to know whether (1) GPT can explain connections between indirectly related entities, (2) create SPARQL queries over the given model and (3) reconstruct the model if all properties and classes were relabelled.

\begin{listing*}
\begin{minted}[autogobble, linenos]{Turtle}
:anne a foaf:Person ; foaf:firstName "Anne" ; foaf:surname "Miller" ;
  vcard:hasAddress [ a vcard:Home ; vcard:country-name "UK" ] .
:bob a foaf:Person ; foaf:firstName "Bob" ; foaf:surname "Tanner" ;
  vcard:hasAddress [ a vcard:Home ; vcard:country-name "US" ] .
:wonderOrg a org:Organization .
:researchDep a org:OrganizationalUnit ; org:unitOf :wonderOrg ;
  rdfs:label "Research Department" .
:marketingDep a org:OrganizationalUnit ; org:unitOf :wonderOrg ;
  rdfs:label "Marketing Department" .
:chiefResearchOfficer a org:Role . :marketingManager a org:Role .
[ a org:Membership ; org:member :anne ; org:organization :researchDep ;
  org:role :chiefResearchOfficer ] .
[ a org:Membership ; org:member :bob  ; org:organization :marketingDep ;
  org:role :marketingManager ] .
\end{minted}
\caption{An organizational KG with two people working in different departments of the same organization.}
\label{lst:example-kg}
\end{listing*}

We issued the following prompt, which includes the knowledge graph from \Cref{lst:example-kg}, on ChatGPT-3 and ChatGPT-4: 
\begin{prompt}
    Given the RDF/Turtle model below, are there any connections between US and UK?   
    $<$rdf-model$>$
\end{prompt}

In the knowledge graph of \Cref{lst:example-kg}, there is a connection between the two countries via the two people living in these, which got a job in different departments of the same company.
While ChatGPT-3 fails to identify this relation, ChatGPT-4 successfully identifies it in all cases.

We further asked both ChatGPT models with prompt \ref{prompt:example-kg-query} and received five SPARQL queries each, which we analysed for their syntactic correctness, plausible query structure, and result quality.
The results for prompt \ref{prompt:example-kg-query} are listed in \cref{tab:vcardSparql} and show that both models produce syntactically correct queries, which in most cases are plausible and produce corrects results in 3/5 (ChatGPT3) and 2/5 (ChatGPT4) cases.

\begin{prompt} \label{prompt:example-kg-query}
    Given the RDF/Turtle model below, create a SPARQL query that lists for every person the country, company and department and role. Please adhere strictly to the given model. 
    $<$rdf-model$>$
\end{prompt}

\begin{table}[tb]
  \caption{Findings in generated SPARQL queries for prompt~\ref{prompt:example-kg-query}.}
  \label{tab:vcardSparql}
  \begin{tabularx}{\textwidth}{ X c c }
    & ChatGPT-3 & ChatGPT-4 \\
    \hline
    syntactically correct & 5/5 & 5/5 \\
    plausible query structure & 4/5 & 3/5 \\
    producing correct result & 3/5 & 2/5 \\
    \hline
    using only defined classes and properties & 3/5 & 4/5 \\
    correct usage of classes and properties & 5/5 & 5/5 \\
    correct prefix for the graph & 5/5 & 4/5 \\
    \end{tabularx}
\end{table}

In essence, AI-based query generation is possible and it can produce valid queries. 
However, the process needs result validation in two dimensions: 1) validating the query itself by matching to static information, like available classes and properties in the graph, as well as 2) validating the executed query results to let ChatGPT generate new queries in case of empty result sets in order to find working queries in a try \& error approach.

As a last prompt on the knowledge graph from \Cref{lst:example-kg}, we created a derived RDF graph by relabelling all classes and properties with sequentially numbered IRIs in the example namespace, like \emph{eg:prop1} and \emph{eg:class2}.
Given the relabelled model, we tasked ChatGPT:
\begin{prompt}
Given the RDF/Turtle model below, please replace all properties and classes with the most likely standard ones.
$<$rdf-model$>$
\end{prompt}

With ChatGPT-3 only 2/5 iterations succeeded in carrying out all substitutions.
In those succeeding cases, the quality was still not as expected because of limited ontology reuse:
Only IRIs in the example namespace were introduced, rather than reusing the \emph{foaf}, \emph{vcard}, and \emph{org} vocabularies.
Yet, the ad-hoc properties and classes were reasonably named, such as \emph{eg:firstName}, \emph{eg:countryName} or \emph{eg:departmentName}.
In contrast, ChatGPT-4 delivered better results: All classes and properties were substituted with those from standard vocabularies - foaf, vcard, and org were correctly identified. 
For some iterations, ChatGPT-4 used the schema.org vocabulary instead of the org vocabulary as an alternative approach.

\subsection{Token Counts for Knowledge Graphs Schemas}
\label{sec:ex:token-counts}
After the results with the small custom knowledge graph we wanted to check the size of some well known knowledge graphs with respect to LLMs.

The LLMs behind ChatGPT can handle at the moment only 4096 tokens (GPT-3.5\textsuperscript{\ref{fn:gpt-3-5-doc}}) or 8192 respective 32,768 tokens for GPT-4\textsuperscript{\ref{fn:gpt-4-doc}}.
 
We counted tokens for various public knowledge graphs in different serialization formats with the library \emph{tiktoken}\footnote{\url{https://github.com/openai/tiktoken}} as recommended for ChatGPT.
\Cref{tab:tokenCounts} lists the token counts for a couple of combinations ordered by token count.
More data and information is available in the \nameref{sec:onlineressources}.
The turtle serialization seem to result in minimal token count, but is still bigger than the similar SQL schema added for comparison.
 All knowledge graphs exceed the token limit for GPT-3.5 and 3 of 4 knowledge graphs listed here exceed the limit for GPT-4.

\begin{table}[tb]
  \caption{Token counts for selected knowledge graphs and serialisations}
  \label{tab:tokenCounts}
  \begin{tabularx}{\textwidth}{ l c r r }
    Graph & Serialisation Type & Token Count \\
    \hline
    Mondial Oracle DB schema & SQL schema & 2,608 token\\
    Mondial RDF schema & turtle & 5,339 token \\
    Mondial RDF schema & functional syntax & 9,696 token \\
    Mondial RDF schema & manchester syntax & 11,336 token\\
    Mondial RDF schema & xml/rdf & 17,179 token \\
    Mondial RDF schema & json-ld & 47,229 token \\
    Wine Ontology & turtle & 13,591 token \\
    Wine Ontology & xml/rdf & 24,217 token \\
    Pizza Ontology & turtle & 5.431 token \\
    Pizza Ontology & xml/rdf & 35,331 token \\
    DBpedia RDF schema & turtle & 471,251 token \\
    DBpedia RDF schema & xml/rdf & 2,338,484 token \\
  \end{tabularx}
\end{table}

\subsection{SPARQL Query Generation for the Mondial Knowledge Graph}
\label{sec:ex:sparql-mondial}
In addition to the experiments with the small custom knowledge graph (see \Cref{sec:ex:sparql-small-custom}) we tested ChatGPT with the bigger mondial knowledge graph\footnote{\url{https://www.dbis.informatik.uni-goettingen.de/Mondial}} which is published since decades with the latest "main revision" 2015.

We asked ChatGPT to generate a SPARQL query for a natural language question from a sparql university lecture\footnote{\url{https://www.dbis.informatik.uni-goettingen.de/Teaching/SWPr-SS20/swpr-1.pdf}}.
We used the following prompt five times with ChatGPT-3 and ChatGPT-4 each:
\begin{prompt} \label{prompt:MondialSparql1Riparian}
Please create a sparql query based on the mondial knowledge graph for the following question: which river has the most riparian states?
\end{prompt}

The results are documented in the \nameref{sec:onlineressources} together with  detailed comments on the given queries.
\Cref{tab:MondialSparql1Riparian} gives some statistics.
In summary, all SPARQL queries given by ChatGPT were syntactically correct, but none of them worked when executed.
Actually all queries had at least one error preventing the correct execution like referencing a wrong namespace, wrong usage of properties or referencing undefined classes.

\begin{table}[tb]
  \caption{Findings in generated sparql queries for prompt~\ref{prompt:MondialSparql1Riparian}.}
  \label{tab:MondialSparql1Riparian}
  \begin{tabularx}{\textwidth}{ X c c }
    & ChatGPT-3 & ChatGPT-4 \\
    \hline
    syntactically correct & 5/5 & 5/5 \\
    plausible query structure & 2/5 & 4/5 \\
    producing correct result & 0/5 & 0/5 \\
    \hline
    using only defined classes and properties & 1/5 & 3/5 \\
    correct usage of classes and properties & 0/5 & 3/5 \\
    correct prefix for mondial graph & 0/5 & 1/5 \\
    \end{tabularx}
\end{table}

\subsection{Knowledge Extraction from Fact Sheets}
\label{sec:ex:fact-sheets-KG}
As an experiment to evaluate knowledge extraction capabilities, we used PDF fact sheets of 3D printer specifications from different additive manufacturing (AM) vendor websites. 
The goal is to build a KG about existing 3D printers and their type as well as capabilities.
We fed plaintext excerpts (extracted via pdfplumber) from these PDFs into ChatGPT-3 and prompted it to:
\begin{prompt}
\label{p:printerExtract}
Convert the following \$\$vendor\$\$ 3d printer specification into a JSON\_LD formatted Knowledge Graph. The node for this KG should be Printer as a main node, Type of 3d printer such as FDM, SLA, and SLS, Manufacturer, Material, Applications, and Technique.
\end{prompt}
Since the fact sheets are usually formatted using a table scheme, the nature of these plain texts is that mostly the printer entity is mentioned in the beginning of the text which then is further characterized in a key-value style. 
As a result, the text typically does not use full sentences and contains only one entity that is described in detail, but several dependant entities (like printing materials).
However, the format of the key-value pairs can be noisy.
Key names can be separated with colons, new line feeds, or in contrast multiple key-value pairs can be in the same line, which could impose a challenge.
Nevertheless, ChatGPT was able to identify the key-value pairs of the evaluation document in a reliably way. 
Unfortunately, it delivered out of 5 test runs for this document 4 partial and 1 complete JSON document.
In spite of that, we summarize first insights gained from a knowledge engineering perspective (but for the sake of brevity, we refer to the output documents in the experiment supplements)
\begin{compactitem}
    \item The JSON-LD output format prioritizes usage of schema.org vocabulary in the 5 evaluation runs. 
    This works good for well-known entities and properties (e.g. \texttt{Organization} @type for the manufacturer, or the \texttt{name} property), however, for the AM-specific feature key names or terms like \texttt{printer} ChatGPT-3 invents reasonable but non-existent property names (in the schema.org namespace) instead of accurately creating a new namespace or using a dedicated AM ontology for that purpose.
    \item Requesting \texttt{turtle} as output format instead, leads to different results. E.g. the property namespace prefix is based on the printer ID and therefore printer descriptions are not interoperable and can not be queried in unified way in a joint KG.
    \item Successfully splitting x,y and z values of the maximum print dimension (instead of extracting all dimensions into one string literal) works in 3 runs. Although ChatGPT-3 accurately appends the unit of measurement to all x,y,z values (which is only mentioned after the z value in the input) in those cases, this is a modelling flaw, as querying the KG will be more complex. In one run it addressed this issue by separating units into a separate unit code field.
    \item A similar effect was observed when it comes to modelling the dependent entities. 
    E.g., in 4 runs, the manufacturer was modelled correctly as a separate typed entity, in 1 as string literal instead.
\end{compactitem}

As a general conclusion of the experiment, ChatGPT-3 has overall solid skills to extract the key value pairs from the sheets, but the correct modelling or representation in terms of a KG significantly varies from run to run.
Subsequently, none of the generated JSON documents contained sufficient information on their own, but only a subset that was modelled accurately. 
A question for future research is whether cherrypicking of individual JSON elements from outputs of several runs and combining them into one final document or iteratively refining the output by giving ChatGPT generic modelling feedback (like use an ontology, or separate unit information, etc.) can be automated in a good and scalable way.

\subsection{Knowledge Graph Exploration}
\label{sec:ex:KG-exploration}
Experts in the field of knowledge graphs are familiar with concepts from RDF Schema (RDFS) (domain/range, subPropertyOf, subClassOf) and Web Ontology Language (OWL) (ObjectProperty, Datatype\-Property, Functional\-Property, ...).
Often, each of these experts has their preferred tools and methods for gaining an overview of an ontology they are not yet familiar with.
We asked ChatGPT-3 two different questions requesting the mermaid\footnote{“… a JavaScript-based diagramming and charting tool …” \url{https://mermaid.js.org/}} visualization of the most important concepts and their connections:

\begin{prompt}
\label{prompt:TopDBpedia}
Can you create me a visualization showing the most important classes and concepts and how they are linked for dbpedia ontology, serialized for mermaid?
\end{prompt}

\begin{prompt}
\label{prompt:TopDBpediaDomainRange}
Can you create me a visualization of the most common concepts of the DBpedia ontology and their connections focusing on domain and range defined in properties.
\end{prompt}

We expected a graph with at least eight nodes and their corresponding edges.
The identifiers for the nodes and edges are expected to follow the Turtle or SPARQL \texttt{prefix:concept} notation.
If the first question did not achieve the goal, we asked additional questions or demands to ChatGPT-3.
The results are presented in \cref{tab:dbpedia} and we evaluated the displayed graphs based on the following criteria:

Prompt \ref{prompt:TopDBpedia} led to an answer with a hierarchical graph representation of the important classes defined in the DBpedia ontology.
The diagram already met our requirements regarding the minimum number and labelling after the first answer and can be seen in the \nameref{sec:onlineressources}.

The class hierarchy was represented by the \texttt{rdfs:subPropertyOf} relation, and the nodes were labelled in prefix notation, as were the edges.
By arranging it as a tree using the subClassOf-pattern, only two different properties were used for the relations (edges). The root node was of type \texttt{owl:Thing} other nodes are connected as (sub)classes from the DBpedia ontology.
These were: Place, Organization, Event, Work, Species, and Person.
The class Work had one more subClassOf relation to the class MusicalWork.
The class Person had the most complex representation, with two more subClassOf relations leading to \texttt{foaf:Person} and \texttt{foaf:Agent}, the latter of which is a subclass of the root node (\texttt{owl:Thing}).

In the second prompt (Prompt \ref{prompt:TopDBpediaDomainRange} ChatGPT-3 referred to a graphic file within the answer text that no longer existed.
Upon further inquiry, a mermaid diagram was generated.
It was of type "Graph" and contained thirteen common concepts and seventeen edges, which were all unique.
The labels of both, nodes and edges contain no prefixes, but were addable with further inquiry.
Only the generated concept \texttt{dbo:Occupation} is non-existent.
All remaining nodes and edges comply with the rules of the ontology, even if the concepts used are derived through further subclass relationships.
The resulting diagram is shown in the \nameref{sec:onlineressources}.

\begin{table}[tb]
  \caption{Diagram content overview.}
  \label{tab:dbpedia}
  \begin{threeparttable}
    
    \begin{tabularx}{\textwidth}{ X c c }
        & Prompt \ref{prompt:TopDBpedia} & Prompt \ref{prompt:TopDBpediaDomainRange} \\
        \hline
        Mermaid Type & graph & graph\tnote{*} \\
        Labels of Nodes & prefix and concept & prefix and concept\tnote{**} \\
        Labels of Edges & prefix and concept & prefix and concept\tnote{**} \\
        Number of Nodes (total/existing/dbo) & 10/10/8 & 13/12/12 \\
        Number of Edges (total/unique) & 12/2 & 17/17 \\
        \hline

    \end{tabularx}
    \begin{tablenotes}
        \item [*] One more prompt was needed to serialize a graph
        \item [**] One more prompt was needed to add prefixed labels
    \end{tablenotes}
  \end{threeparttable}
\end{table}

While prompt \ref{prompt:TopDBpedia} leads to a result that can be more comprehensively achieved with conventional tools for visualizing RDF, the result from prompt \ref{prompt:TopDBpediaDomainRange} provides an overview of concepts (classes) and properties that can be used to relate instances of these classes to each other.

\section{Conclusion and Future Work}

From the perspective of a knowledge graph engineer, ChatGPT has demonstrated impressive capabilities.
It successfully generated knowledge graphs from semi-structured textual data, translated natural language questions into syntactically correct and well-structured SPARQL queries for the given knowledge graphs, and even generated overview diagrams for large knowledge graph schemas, as outlined in \cref{sec:experiments}.
An detailed analysis revealed that the generated results contain mistakes, of which some are subtle.
For some use cases, this might be harmless and can be tackled with additional validation steps in general, like with the metrics we used for SPARQL queries.
In general, our conclusion is, that one needs to keep in mind ChatGPT's tendency to \emph{hallucinate}\footnote{Generation of content without any foundation}, especially when applied to the field of knowledge graph engineering where many engineers are used to mathematical precision and logic.

The closed-source nature of ChatGPT challenges scientific research on it in two ways:
\begin{inparaenum}
    \item Detailed capability ratings of closed-source probabilistic models require much effort
    \item Result reproducibility is bound to service availability and results might be irreproducible at a later date (due to service changes)
\end{inparaenum}
Thus, open training corpora and LLMS are mandatory for proper scientific research.

In the future, metrics are to be found to rate generated ChatGPT answers automatically, like we broached with SPARQL queries. This again enables to extend the number of test cases for a specific experiment and to generate profound statistical results. Another research focus should be given to methods that let the LLM access a broader/necessary context to increase the chance for correct answers.

\subsubsection*{Acknowledgements} 
This work was partially supported by grants from the German Federal Ministry for Economic Affairs and Climate Action (BMWK) to the CoyPu project (01MK21007A) and KISS project (01MK22001A) as well as from the German Federal Ministry of Education and Research (BMBF) to the project StahlDigital (13XP5116B) and project KupferDigital (13XP5119F).

\vspace{10\baselineskip}

%
%
\bibliographystyle{splncs04}
\bibliography{ref.bib}

\appendix
\section*{Supplemental Online Resources}\label{sec:onlineressources} 
Details on the experiments described can be found at the following github repo: \url{https://github.com/AKSW/AI-Tomorrow-2023-KG-ChatGPT-Experiments}

\newpage
\section{Deutsche Zusammenfassung}

Wissensgraphen (englisch \emph{Knowledge Graphs}, KGs), bieten uns eine strukturierte, flexible, transparente, systemübergreifende und kollaborative Möglichkeit, unser Wissen und unsere Daten über verschiedene Bereiche der Gesellschaft und der industriellen sowie wissenschaftlichen Disziplinen hinweg zu organisieren. KGs übertreffen jede andere Form der Repräsentation in Bezug auf die Effektivität. Die Entwicklung von Wissensgraphen (englisch \emph{Knowledge Graph Engineering}, KGE) erfordert jedoch fundierte Erfahrungen mit Graphstrukturen, Webtechnologien, bestehenden Modellen und Vokabularen, Regelwerken, Logik sowie Best Practices. Es erfordert auch einen erheblichen Arbeitsaufwand.

In Anbetracht der Fortschritte bei großen Sprachmodellen (englisch \emph{Large Language Modells}, LLMs) und ihren Schnittstellen und Anwendungen in den letzten Jahren haben wir umfassende Experimente mit ChatGPT durchgeführt, um sein Potenzial zur Unterstützung von KGE zu untersuchen. In diesem Artikel stellen wir eine Auswahl dieser Experimente und ihre Ergebnisse vor, um zu zeigen, wie ChatGPT uns bei der Entwicklung und Verwaltung von KGs unterstützen kann.

\end{document}